\documentclass[10pt,twocolumn,letterpaper]{article}

\usepackage{iccv}
\usepackage{times}
\usepackage{epsfig}
\usepackage{graphicx}
\usepackage{amsmath}
\usepackage{amssymb}
\usepackage{multirow}
\usepackage{booktabs}
\usepackage{pifont}
\usepackage{marvosym}
\usepackage{hyperref}
\hypersetup{
    colorlinks=true,
    linkcolor=blue
    }

\iccvfinalcopy 

\ificcvfinal\fi

\newcommand{\fakepara}[1]{\vspace{1mm}\noindent\textbf{#1.}}
\DeclareMathAlphabet{\pazocal}{OMS}{zplm}{m}{n}

\newcommand{\cmark}{\ding{51}}%
\newcommand{\xmark}{\ding{55}}%

\begin{document}

\title{NeILF++: Inter-Reflectable Light Fields for Geometry and Material Estimation}
\author{Jingyang Zhang \quad Yao Yao\textsuperscript{\Letter} \quad Shiwei Li \quad Jingbo Liu \quad Tian Fang \\ David McKinnon \quad Yanghai Tsin \quad Long Quan \\
Apple \\
{\tt\small \{jingyang\_zhang, yaoyao, shiwei, jingbo, fangtian, dmckinnon, ytsin, quan\_long\}@apple.com}
}

\maketitle

\vspace{-2mm}

\begin{abstract}


We present a novel differentiable rendering framework for joint geometry, material, and lighting estimation from multi-view images. In contrast to previous methods which assume a simplified environment map or co-located flashlights, in this work, we formulate the lighting of a static scene as one neural incident light field (NeILF) and one outgoing neural radiance field (NeRF). The key insight of the proposed method is the union of the incident and outgoing light fields through physically-based rendering and inter-reflections between surfaces, making it possible to disentangle the scene geometry, material, and lighting from image observations in a physically-based manner. The proposed incident light and inter-reflection framework can be easily applied to other NeRF systems. We show that our method can not only decompose the outgoing radiance into incident lights and surface materials, but also serve as a surface refinement module that further improves the reconstruction detail of the neural surface. We demonstrate on several datasets that the proposed method is able to achieve state-of-the-art results in terms of geometry reconstruction quality, material estimation accuracy, and the fidelity of novel view rendering. Project page: \url{https://yoyo000.github.io/NeILF_pp}.

\end{abstract}

\section{Introduction}
\label{sec:intro}


Reconstructing 3D scene information from multi-view images is a persistent challenge in the field of computer vision and computer graphics. The Neural Radiance Field (NeRF) approach~\cite{mildenhall2020nerf}, which uses differentiable volume rendering to jointly optimize scene geometry and appearance, offers a novel solution and has demonstrated great results in novel view synthesis and neural surface reconstruction. While NeRF successfully models the outgoing radiance of a scene, it fails to disentangle the incident lighting and surface properties from the radiance, preventing its use in downstream applications such as object relighting and material property editing.

Recent works have acknowledged the problem and proposed to further decompose the outgoing radiance into the material and environmental lighting. Some of them adopt simplified lighting models to reduce the computational complexity, including co-located flash light~\cite{nam2018practical,bi2020deep3d,bi2020deep,schmitt2020joint} and environment map~\cite{zhang2020physg,zhang2020nerfactor,boss2021nerd,munkberg2021nvdiffrec,boss2021neuralpil}. These models are insufficient to model all kinds of lighting configurations, such as non-distant light sources and inter-reflections between surfaces. Recently, the Neural Incident Light Field (NeILF)~\cite{yao2022neilf} instead models arbitrary static lighting conditions by encoding spatially-varying incident lighting in a neural network. However, it only provides the modeling capability of, but does not introduce an explicit constraint on the inter-reflection issue. Moreover, NeILF requires an object mesh as input, whose quality has a strong influence on the estimated material.

\begin{figure}
	\centering
	\includegraphics[width=1\linewidth]{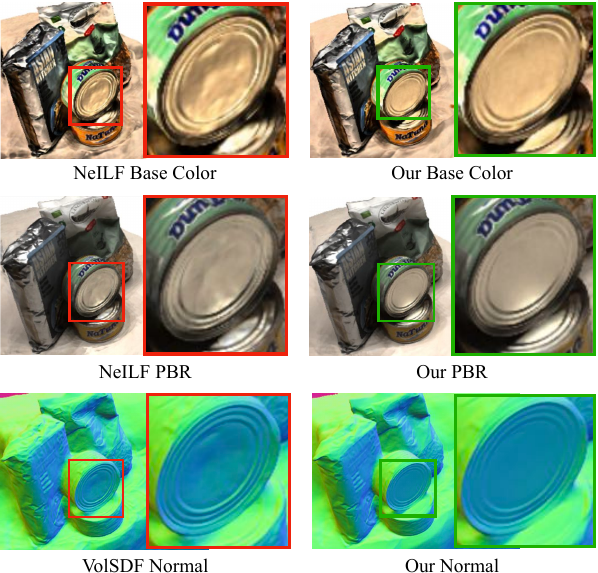}
	\caption{Compared with the baseline that simply runs NeILF \cite{yao2022neilf} following VolSDF \cite{yariv2020multiview}, the proposed approach is able to jointly optimize the scene geometry and material. The neural surface, material, and final rendering quality are improved.}
	\label{fig:teaser}
\end{figure}

In this work, we demonstrate that suitable lighting representation is the key to joint geometry and material estimation from multi-view images. Without loss of generality, the lighting of a typical scene can be represented by the incident light to the object surface and the outgoing radiance from the surface. Based on the observation, we propose to model the light fields, geometry, and materials of the scene as four separated fields, namely 1) one outgoing radiance field \cite{mildenhall2020nerf}, 2) one incident light field \cite{yao2022neilf}, 3) one signed distance field as the scene geometry, and 4) one field of bidirectional reflectance distribution function (BRDF) parameters as the surface material. 
The key insight of the proposed method is that the incident light and the outgoing radiance can be naturally unified through physically-based rendering and inter-reflections between surfaces. We demonstrate that with the union of the outgoing radiance and the incident light field, the proposed method not only can generate high-quality BRDF estimation for object relighting, but also improves the reconstruction detail of the surface geometry. The proposed approach can be easily applied to different NeRF systems for material decomposition and surface detail refinement. 



To better evaluate the quality of estimated material from physically-based rendering, we construct a real-world linear high-dynamic range (HDR) dataset called NeILF-HDR. With this, we can avoid adding an HDR-LDR conversion module and directly train in HDR color space. We have extensively studied our approach on the DTU dataset~\cite{jensen2014large}, the NeILF synthetic dataset~\cite{yao2022neilf}, and the NeILF-HDR dataset. We show that the proposed method is able to achieve state-of-the-art results in geometry reconstruction quality, material estimation accuracy, and novel view rendering quality. To summarize, major contributions of this paper include:

\begin{itemize}
	\item Proposing a general light field representation by marrying one incident light field and one outgoing radiance field via PBR and inter-reflections.
	\item Proposing an optimization scheme for joint geometry, material, and lighting estimation, which can be easily applied to the prevalent NeRF family for material decomposition and neural surface refinement.
	\item Constructing a real-world linear HDR dataset for material estimation and other neural rendering tasks.  
\end{itemize}

\begin{figure*}
	\centering
	\includegraphics[width=1\linewidth]{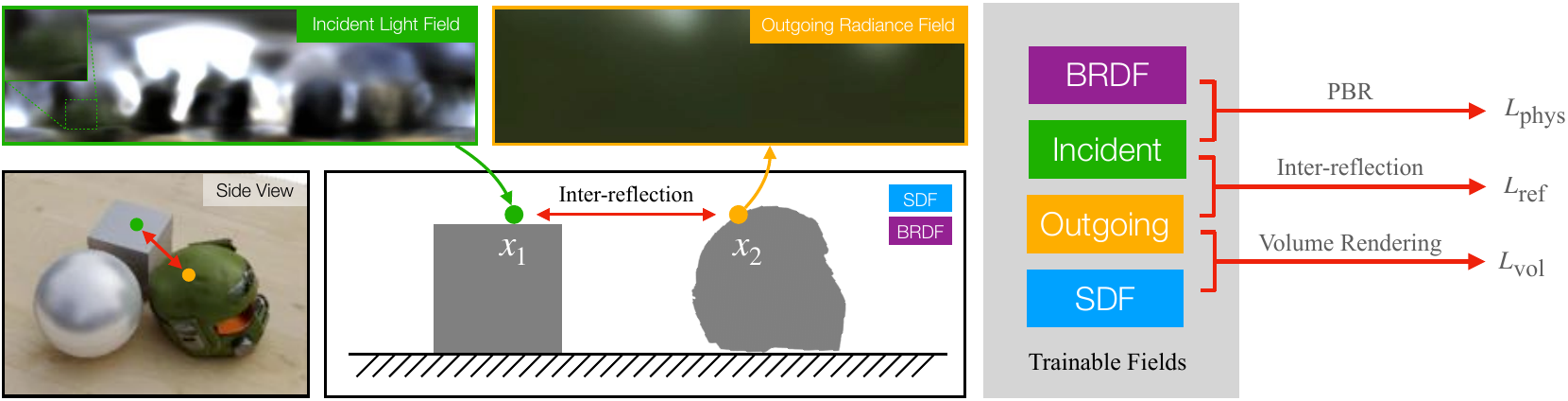}
	\caption{Illustration on the proposed light field representation. The incident light and the outgoing light are constrained by physically-based rendering (as in NeILF) and inter-reflection between surfaces. The inter-reflection constraint ensures the consistency between the two lighting representations during the joint optimization of the four fields. The losses $L_\text{phys}$, $L_\text{vol}$ and $L_\text{ref}$ are for the physically rendered image, the volume rendered image, and the inter-reflection constraint respectively, which will be elaborated in Sec.~\ref{sec:method-joint}. }
	\label{fig:lightfields}
\end{figure*}


\section{Related Works}
\label{sec:related}

\fakepara{The Rendering Equation} With the given BRDF properties of a surface point $\mathbf{x}$, the outgoing light from $\mathbf{x}$ can be computed via the incident light to the surface~\cite{kajiya1986rendering}:
\begin{equation}\label{eq:rendering}
	\begin{aligned}
		L_o^P(\boldsymbol{\omega_o}, \mathbf{x}) &= \int_{\Omega} f(\boldsymbol{\omega_o}, \boldsymbol{\omega_i}, \mathbf{x}) L_i(\boldsymbol\omega_i, \mathbf{x}) (\boldsymbol\omega_i\cdot \mathbf{n}) d\boldsymbol\omega_i,
	\end{aligned}
\end{equation}
where $\boldsymbol\omega_o$ is the viewing direction of the outgoing light, $\mathbf{n}$ is the surface normal, $ L_i $ is the incident light from direction $\boldsymbol\omega_i$, and $f$ is the BRDF properties of the surface point. 

The goal of joint geometry, material, and lighting estimation is to recover the surface location $\mathbf{x}$, the scene lighting $L_i$ and the BRDF function $f$ in the above equation. Below we give a brief review of the physically-based optimization approaches with the multi-view image inputs.

\fakepara{Lighting Modeling for Material Estimation}
Due to the high-dimensional nature of lighting and material, it is rather difficult to jointly recover both from multi-view image observations. Previous methods usually apply controlled lighting to simplify the problem. For example, point/co-located flash lights~\cite{nam2018practical,bi2020deep3d,bi2020deep,schmitt2020joint,li2022neural}, turn-table models~\cite{dong2014appearance,xia2016recovering} and the environment map assumption~\cite{zhang2020physg,zhang2020nerfactor,boss2021nerd} are applied to reduce the complexity of the problem. Also, some works apply additional sensors~\cite{guo2019relightables,azinovic2019inverse,park2020seeing} to facilitate the optimization process.
However, such mitigations will inevitably limit these methods to real-world applications. 

Other works instead adopt more complex lighting models for material and lighting estimation. NeRV \cite{srinivasan2021nerv} proposes to use reflectance and visibility fields to model the indirect light. However, it requires the environment map to be given in advance and its pipeline is computationally expensive. ShadowNeuS \cite{ling2022shadowneus} and NeRFactor \cite{zhang2020nerfactor} exploit light visibility for material and incoming light decomposition, but it fails to model the inter-reflections between surfaces. A joint illumination and material
estimation approach is proposed by Nimier-David et al.~\cite{nimier2021material}, but requires the given geometry and costly multi-bounce raytracing. Recently, NeILF \cite{yao2022neilf} is proposed to model arbitrary static lighting by using a spatially-varying incident light field. However, NeILF also requires the geometry to be given and its incident light setting does not explicitly take inter-reflections into account. In this work, we propose to model both incident and outgoing lights of the surface, where the two light fields are further unified through PBR and inter-reflections. 

\fakepara{Surface Optimization by Differentiable Rendering}
Recent neural rendering methods have shown promising results in scene geometry recovery. The scene geometry is usually represented by an implicit function, such as a density field~\cite{mildenhall2020nerf,zhang2020nerf++,zhang2020nerfactor,boss2021nerd}, an occupancy field~\cite{niemeyer2020differentiable}, or a signed distance field~\cite{yariv2020multiview,yariv2021volume,wang2021neus,zhang2021learning,zhang2022critical}. During the differentiable rendering optimization, both the scene geometry and appearance will be recovered by minimizing the difference between rendered images and input images. 

Recent neural surface reconstruction methods can be categorized into 1) differentiable surface rendering and 2) differentiable volume rendering approaches. To render a pixel, differentiable surface rendering will first find the surface intersection using root-finding algorithms and then query the color of the intersection. To make the process differentiable, the first-order approximation of the surface intersection is usually applied~\cite{niemeyer2020differentiable,yariv2020multiview}. The method is fast, however, can only refine the geometry near the surface, and often requires additional geometric priors (e.g., silhouette~\cite{yariv2020multiview,zhang2020physg}, depth~\cite{zhang2021learning} or point cloud~\cite{zhang2022critical}) for robust estimation. The second type of method uses volume rendering to accumulate all the radiance along the ray weighted by the density and visibility of matter \cite{mildenhall2020nerf,wang2021neus,yariv2021volume}. The method is good at recovering the topology and has gained much popularity recently due to its simplicity. 
In this work, we show that the surface detail from a volume rendering system can be further refined by combining the differentiable physically-based rendering system (i.e., NeILF).


\section{Inter-reflectable Light Fields}
\label{sec:method-ref}

\subsection{Recap on Incident Light and BRDF Modeling}

\fakepara{Neural Incident Light Field}\label{sec:neilf}
As proposed by NeILF~\cite{yao2022neilf}, the incoming lights in the scene can be formulated as a neural incident light field, 
which is recorded by a multi-layer perceptron (MLP). The MLP takes a point location $\mathbf{x}$ and a direction $\boldsymbol\omega$ as inputs, and returns an incident light $L$:
\begin{equation}
	\begin{aligned}
		\mathbb{L}:  \{\mathbf{x}, \boldsymbol\omega_i\} \to \mathbf{L}.
	\end{aligned}
\end{equation}

 Compared with environment maps, the specially-varying illumination given by incident light fields is capable of modeling direct/indirect light and occlusions of any static scenes, and thus facilitates the material estimation in scenes with such complex geometry and lighting condition. 

\fakepara{BRDF parameterization}
We use a simplified Disney principled BRDF parameterization as in \cite{yao2022neilf}. For a surface point $\mathbf{x}$, the BRDF field $ \mathbb{B}: \mathbf{x} \to \{\mathbf{b}, r, m\} $ stores a base color $ \mathbf{b} \in [0,1]^3 $, a roughness $ r \in [0,1] $ and a metallic $ m \in [0,1] $. The BRDF $f$ in Eq.~\ref{eq:rendering} is the summation of a diffuse term $f_d$ and a specular term $f_s$. The diffuse term can be calculated as $f_d = \frac{1 - m}{\pi} \cdot \mathbf{b}$, and the specular term can be computed as:
\begin{equation}\label{eq:f_s}
	\begin{aligned}
		f_s(\boldsymbol\omega_o, \boldsymbol\omega_i) &= \frac{D(\mathbf{h}; r) \cdot F(\boldsymbol\omega_o, \mathbf{h}; \mathbf{b}, m) \cdot G(\boldsymbol{\omega}_i, \boldsymbol\omega_o, \mathbf{h}; r)}{4 \cdot (\mathbf{n} \cdot \boldsymbol\omega_i) \cdot (\mathbf{n} \cdot \boldsymbol\omega_o)}, 
	\end{aligned}
\end{equation}
where $\mathbf{h}$ is the half vector between the incident direction $\boldsymbol\omega_i$ and the viewing direction $\boldsymbol\omega_o$. D, F, and G refer to the normal distribution function, the Fresnel term, and the geometry term respectively. We adopt similar implementation of D, F, and G as in previous works~\cite{wang2009all,zhang2020physg,yao2022neilf} and details are provided in the supplementary material.


\subsection{Unifying Incident and Outgoing Lights with Inter-reflection}

NeILF represents the incoming light to the surface but does not explicitly model the outgoing light. On the other hand, as reported in previous works, the outgoing light can be efficiently recovered in differentiable rendering frameworks. For example, NeRF~\cite{martin2021nerf} recovers the radiance in space, while the implicit differentiable renderer (IDR~\cite{yariv2020multiview}) optimizes the surface light field of an object. To model a complete light field, we propose to additionally add one outgoing radiance field to the NeILF framework. 

A straightforward solution to recover the proposed light fields is to optimize a NeILF~\cite{yao2022neilf} system and an IDR~\cite{yariv2020multiview} system independently. However, such an optimization strategy would lead to inconsistent incident and outgoing lights. In fact, the two light fields can be unified through inter-reflections between surfaces. As illustrated in Fig.~\ref{fig:lightfields}, assuming that the space between object surfaces is empty, the outgoing light from $\textbf{x}_2$ in the helmet to $\textbf{x}_1$ should be equal to the incident light to $\textbf{x}_1$ in the cube.

Implementation-wise, the incident and outgoing light is constrained by a consistency loss $L_{\text{ref}}$. As in NeILF, we sample incident rays and query their intensity during PBR. Then, we trace the reversed ray and check whether it hit any surface. If a ray $ -\boldsymbol{\omega_i} $ from $ \mathbf{x}_1 $ hit the surface at $ \mathbf{x}_2 $, we query the intensity from the outgoing radiance field $\mathbb{R}$ and encourage the two light to be the same. 
\begin{equation}
	L_{ref} = \| \mathbb{L}(\mathbf{x}_1, \boldsymbol{\omega_i}) - \mathbb{R}(\mathbf{x}_2, \boldsymbol{\omega_i}) \|_1.
\end{equation}
Compared to a hard constraint that directly uses the queried outgoing light as the incident light, we do not need to trace a large amount of all incident rays. This can greatly reduce the computational cost as ray tracing is time-consuming especially when volume rendering is applied (Sec.~\ref{sec:method-joint}). 

\section{Joint Geometry Optimization}
\label{sec:method-joint}

In this section, we describe how the proposed inter-reflectable light fields can be used for geometry and material estimation. We first discuss our geometry representation in Sec.~\ref{sec:sdf}, and then describe details of the proposed optimization scheme in Sec.~\ref{sec:joint}.

\subsection{Recap on Geometry Representation}\label{sec:sdf}

\fakepara{Signed Distance Field}
Following recent neural surface reconstruction approaches \cite{yariv2020multiview,wang2021neus,yariv2021volume,zhang2021learning,zhang2022critical}, we represent the scene geometry as a signed distance field (SDF) in space. The surface $ \pazocal{S} $ is the zero level set of the SDF $ \mathbb{G} $ represented by a neural network with parameters $\theta$. Let $ \Omega \subset \mathbb{R}^3 $ be the domain of 3D space, the network takes a point location $ \mathbf{x} $ as input and outputs its distance to the nearest surface point: $ \mathbb{G}: \Omega \rightarrow \mathbb{R} $. The surface $ \pazocal{S} $ is defined as:
\begin{equation} \label{eq:levelset}
	\pazocal{S} = \{ \mathbf{x} \in \Omega \;|\; \mathbb{G}(\mathbf{x};\theta)=0 \}.
\end{equation}

\fakepara{Volume Rendering}
The volume rendering technique from NeRF can globally optimize the scene geometry as the alpha composition process is dependent on points in the whole space rather than only on the surface. Let the outgoing radiance field be $ \mathbb{R}(\mathbf{x}_i, \boldsymbol\omega_o) $, we render $ L_o^R $ by alpha-blending the samples $ \{ \mathbf{x}_i \}_{i=1}^N $ along the ray with direction $ -\boldsymbol\omega_o $: 
\begin{equation}
\begin{aligned}
	L_o^R &= \sum_{i=1}^{N} T_i (1 - \exp (-\sigma_i \delta_i) ) \mathbb{R}(\mathbf{x}_i, \boldsymbol\omega_o),  \\ 
	& \text{where } T_i = \exp( -\sum_{j=1}^{i-1} \sigma_j \delta_j ), 
\end{aligned}
\end{equation}
In addition, recent works \cite{wang2021neus, yariv2021volume} proposed to further transform an SDF to a density field. In this paper, we adopt the transformation from VolSDF \cite{yariv2021volume} for its simplicity:
\begin{equation}
	\sigma_i = \alpha \Psi_{\beta} (- \mathbb{G}(\mathbf{x}_i) ), 
\end{equation}
where $ \Psi_{\beta} $ is the cumulative distribution function of the Laplace distribution with zero mean and $ \beta $ scale. 


\subsection{Joint Shape, Material and Lighting Estimation}\label{sec:joint}

As discussed in previous sections, a 3D scene is parameterized by a SDF field $ \mathbb{G}(\mathbf{x}) $ as the scene geometry, a BRDF field $ \mathbb{B}(\mathbf{x}) $ as the surface material, an incident light field $ \mathbb{L}(\mathbf{x}, \boldsymbol{\omega_i}) $ and a outgoing radiance field $ \mathbb{R}(\mathbf{x}, \boldsymbol{\omega_o}) $ as the scene lighting. In each iteration of joint training, the losses for joint training (shown in Fig.~\ref{fig:lightfields}) are calculated by the following steps: 
\begin{enumerate}
    \vspace{-2mm}
    \item Perform volume rendering to get the color for a ray from the outgoing radiance field, and supervise it with ground truth ($L_{\text{vol}}$). 
    \vspace{-2mm}
    \item Find the position and normal of the intersection point between the ray and the surface by alpha blending.
    \vspace{-2mm}
    \item Feed the position into the BRDF field to get the material parameters.
    \vspace{-2mm}
    \item Sample incident light directions from the upper hemisphere and query the incident light from the incident light field. 
    \vspace{-2mm}
    \item Feed the BRDF parameters, the surface normal and the incident lights into the rendering equation to get the PBR color, and supervise it with ground truth ($L_{\text{phys}}$). 
    \vspace{-2mm}
    \item Back trace the incident directions, get the color by volume rendering from the outgoing radiance field, and minimize its difference with the incident light ($L_{\text{ref}}$). 
    \vspace{-2mm}
\end{enumerate}
In this procedure, we do not require any specific design from the geometry and the outgoing radiance field. Abstractly, the PBR requires the following output from these two fields for a ray: 1) volume rendered color and 2) information of the intersection point with the surface, including both position and normal. These requirements can be fulfilled by most NeRF-like systems, which guarantees the generalization of our system and the capability for future upgrades.

\begin{figure*}
	\centering
	\includegraphics[width=1\linewidth]{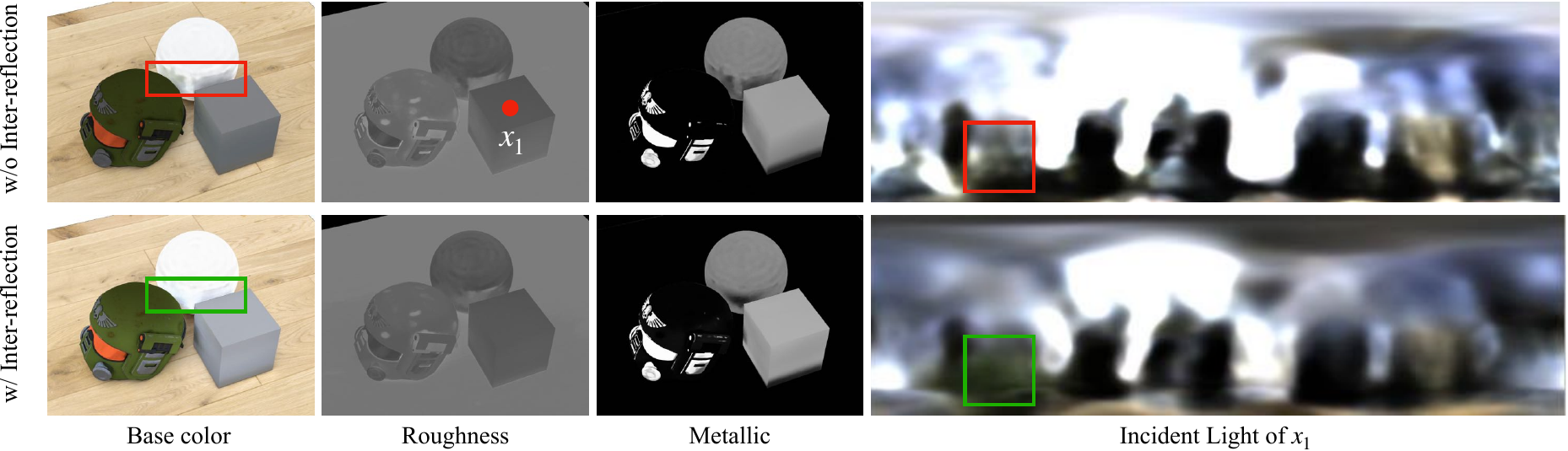}
	\caption{Qualitative ablation study on the inter-reflection constraint. The incident light map of $\mathbf{x}_1$ on the cube successfully matches the outgoing radiance from the helmet. And reflections are not baked into the base color. }
	\label{fig:ref}
\end{figure*}

\begin{table}[t]
	\resizebox{\linewidth}{!}{%
		\begin{tabular}{l|c|ccccc|ccc}
			\specialrule{.2em}{.1em}{.1em}
			\multirow{2}{*}{Stage} & Joint & \multicolumn{5}{c|}{Geometry} & \multicolumn{3}{c}{Material} \\
 & $L_{\text{ref}}$ & $L_{\text{vol}}$ & $L_{\text{Eik}}$ & $L_{\text{Hess}}$ & $L_{\text{surf}}$ & $L_{\text{pcd}}$ & $L_{\text{phys}}$ & $L_{\text{smth}}$ & $L_{\text{Lam}}$ \\ \hline
SDF Init. & \xmark & \cmark & \cmark & \cmark & \cmark & $\blacksquare$ & \xmark & \xmark & \xmark \\
Mat. Init. & \cmark & \xmark & \xmark & \xmark & \xmark & \xmark & \cmark & \cmark & \cmark \\
Joint & \cmark & \cmark & \cmark & $\downarrow$ & $\downarrow$ & \xmark & \cmark & \cmark & \cmark \\
			\specialrule{.2em}{.1em}{.1em}
	\end{tabular}}
	\caption{Used losses in each training stage. $\blacksquare$: The point cloud supervision is optional. $\downarrow$: The weight of this loss is downscaled than the previous stage. }
	\label{tab:losses}
\end{table}

\subsection{Training Scheme}
 To facilitate convergence, we propose the following 3-stage scheme to joint optimize the four fields: 

\fakepara{SDF Initialization}
First we initialize the SDF so that it recovers the basic topology of the object. The training strategy is similar to common NeRF frameworks. We only do step 1 described in Sec.~\ref{sec:joint}. In addition, we use Eikonal loss $L_{\text{Eik}}$ \cite{yariv2020multiview}, Hessian loss $L_{\text{Hess}}$ and minimal surface loss $L_{\text{surf}}$ \cite{zhang2022critical} to regularize the surface. Optionally, an oriented point cloud can be used to directly supervise the surface ($L_{\text{pcd}}$) \cite{zhang2022critical}, which further enhances the stability of the optimization. In this stage, the material estimation module does not participate in the calculation, and thus is not optimized. 

\fakepara{Material Initialization}
In the second stage, we also initialize the material and the lighting to facilitate convergence in the joint training stage. We do steps 2-6 in Sec.~\ref{sec:joint}. In this stage, the intersection point is treated as a constant and the geometry is fixed. Regularization including bilateral smoothness $L_{\text{smth}}$ and Lambertian regularization $L_{\text{Lam}}$ \cite{yao2022neilf} are also added. 

\fakepara{Joint Optimization with Inter-reflections}
In this stage, we jointly optimize all four fields. Now we require the intersection points to be differentiable so that the PBR loss can optimize the geometry through the normal fed into the rendering equation and the position fed into the BRDF and incident light field. The inter-reflection loss is also added as the previous stage. Specially, we reduce some of the geometry smoothness terms to facilitate the recovery of fine details. A comprehensive list of used losses is presented in Tab.~\ref{tab:losses}. 

\subsection{HDR Rendering} \label{sec:hdr}
In this section, we claim that linear HDR input is necessary for both the quality of the estimated material and the correctness of the inter-reflection constraint. Because the common LDR images are usually processed by unknown non-linear tone-mapping, gamma correction, and value clipping, it may result in inaccurate material and lighting estimation if we directly supervise the rendering value with the LDR ground truth. For example, because the observed intensity of a specular area is suppressed or clipped, the intensity of light sources is often underestimated. Converting the raw render to the same color space as the LDR images by either fixed or learned \cite{yao2022neilf} mapping is still problematic because of potential information loss. For example, the saturated pixels are never supervised because the clipping operation eliminates the gradients. 
On the other hand, the outgoing radiance field also needs to be trained in HDR because it is used to regularize the incident light in the inter-reflection constraint. 
Therefore, linear HDR images are necessary for our method. 
In Sec.~\ref{sec:exp-overall}, we show that HDR supervision results in a more robust estimation than LDR. 

\section{Experiments}

Our method has been evaluated on 1) the DTU dataset, 2) an in-house synthetic dataset (NeILF-synthetic) \cite{yao2022neilf} and 3) an in-house captured linear HDR dataset (NeILF-HDR). In Sec.~\ref{sec:exp-reflect}, we do an ablation study on the inter-reflection handling mechanism, and in Sec.~\ref{sec:exp-joint} we compare the effect of joint geometry and material estimation. In Sec.~\ref{sec:exp-overall}, we report the geometry accuracy and fidelity of novel view rendering for real-world datasets.

\begin{table}
	\resizebox{\linewidth}{!}{%
		\begin{tabular}{l|c|ccc|ccc|c}
			\specialrule{.2em}{.1em}{.1em}
			 \multirow{2}{*}{}           & \multirow{2}{*}{$L_{\text{ref}}$} & \multicolumn{3}{c|}{Env} & \multicolumn{3}{c|}{Mix} & \multicolumn{1}{l}{\multirow{2}{*}{Mean}} \\
			 &                        & city  & studio & castel & city  & studio & castel & \multicolumn{1}{l}{}                      \\ \hline
			\multirow{2}{*}{Base Color} & \xmark & 18.09 & 17.75  & 16.33  & \textbf{17.93}     & \textbf{17.26}       & \textbf{18.69}       & 17.68 \\
			& \cmark & \textbf{18.23} & \textbf{19.29}  & \textbf{17.47}  & 17.38     & 17.13       & 18.19       & \textbf{17.95} \\ \hline
			\multirow{2}{*}{Roughness}  & \xmark & 20.34 & 22.20  & 19.69  & 21.31     & \textbf{23.23}       & 21.53       & 21.38 \\
			& \cmark  & \textbf{21.31} & \textbf{22.43}  & \textbf{20.98}  & \textbf{21.79}     & 23.05       & \textbf{21.84}       & \textbf{21.90} \\\hline
			\multirow{2}{*}{Metallic}   & \xmark & 18.56 & 17.13  & 17.09  & 20.52     & 18.18       & \textbf{18.50}       & 18.33 \\
			& \cmark & \textbf{19.42} & \textbf{18.59}  & \textbf{17.26}  & \textbf{20.58}     & \textbf{20.67}       & 17.96       & \textbf{19.08} \\ 
			\specialrule{.2em}{.1em}{.1em}
	\end{tabular}}
	\caption{Quantitative ablation study on the effect of the inter-reflection constraint on material estimation. The proposed inter-reflection improves the overall quality of the estimated material. }
	\label{tab:interref}
\end{table}

\begin{figure*}
	\centering
	\includegraphics[width=1\linewidth]{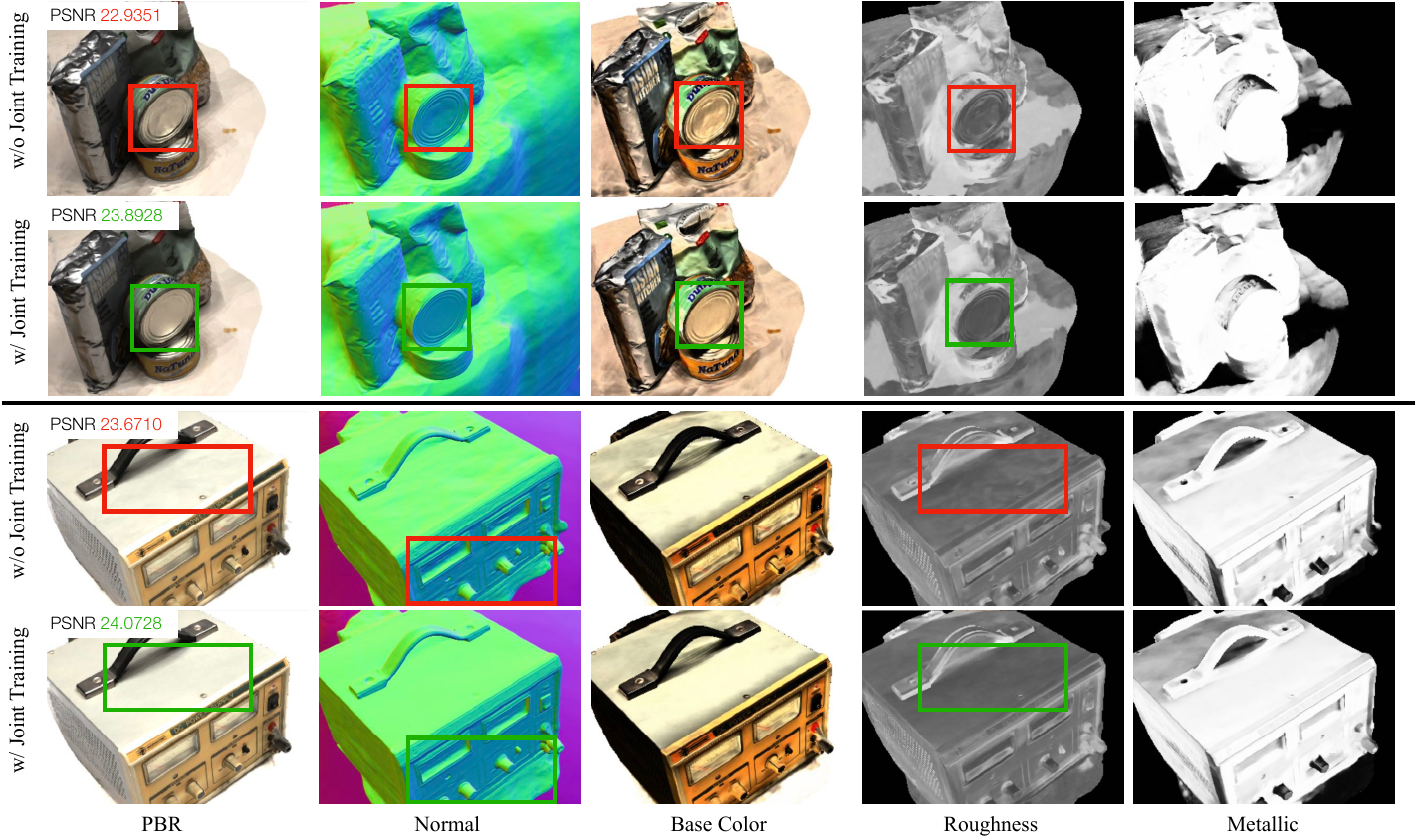}
	\caption{Qualitative ablation study on the joint optimization. For the baseline method, we use a modified VolSDF~\cite{yariv2020multiview} to generate the geometry and use NeILF~\cite{yao2022neilf} to generate the surface material. With the joint refinement, the geometry and material estimation and the final rendering quality are  improved.}
	\label{fig:comparison}
\end{figure*}

\subsection{Implementation}

\fakepara{Network Architecture}
For the network architecture, we test both the heavy MLP with sinusoidal positional encoding (Ours (MLP)) \cite{mildenhall2020nerf} and the lightweight MLP with hash grid encoding (Ours (NGP)) \cite{mueller2022instant}. Because the hash grid based network has weaker intrinsic smoothness, we additionally use a point cloud to facilitate the convergence in the geometry initialization stage, where the point cloud can be automatically generated from off-the-shelf multi-view stereo methods \cite{schonberger2016pixelwise,zhang2020visibility}. Note that point cloud supervision is removed for all architecture in the joint training stage. 

\fakepara{HDR Rendering}
The dynamic range handling of the material estimation module follows NeILF \cite{yao2022neilf}. The raw rendered color is directly supervised by the HDR ground truth. If the input is in LDR, we clip the rendered color to $[0,1]$ and convert to the sRGB color space. The incident field network uses an exponential activation for the last layer for two reasons: 1) It guarantees the non-negativity of light intensity. 2) It regularizes the value distribution of pre-activated output, given that usually the light sources in the environment have extremely large intensity compared with the light from other directions. For the outgoing radiance field, we scale up the output from the Sigmoid activation to support the unbounded color. To preserve the magnitude of the gradient, we scale down the loss of the outgoing radiance field $L_{\text{vol}}$ rendering accordingly.

\fakepara{Time and Memory}
In each iteration, we sample 4096 rays with 64 volume rendering samples. During PBR we sample 128 incident rays for each surface point. We conduct all the experiments on a single Tesla V100 GPU. For the heavy backbone, we train the system for 80000 steps which takes 24 hours and 30GB of VRAM. For the lightweight backbone, we  train the system for 30000 steps which takes 4 hours and 10GB of VRAM. 




\subsection{Evaluation on Inter-reflectable Light Fields}\label{sec:exp-reflect}
We do an ablation study on the inter-reflection constraint. The experiments are conducted on the NeILF-synthetic~\cite{yao2018mvsnet} dataset that contains three objects placed on a plane and lit by 6 different lighting conditions. The dataset provides raw HDR images, geometry, and ground truth material for evaluation. In this experiment, we focus on material estimation so we use the given geometry and keep it fixed. For each scene, we split 9 images from all 96 images as validation views and report the accuracy of BDRF parameters and rendering of these views in terms of peak signal-to-noise ratio (PSNR).

As is shown in Fig.~\ref{fig:ref}, the incident light map at the point $ \mathbf{x}_1 $ on the cube successfully captures the objects beside it. Because lights in the occlusion-severe region are correctly constrained, colors reflected from other objects are not baked into the base color anymore. Quantitative results (Tab.~\ref{tab:interref}) also show the overall improvements of the estimated materials. 

\begin{table*}
	\resizebox{\textwidth}{!}{%
		\begin{tabular}{l|l|ccccccccccccccc|c}
			\specialrule{.2em}{.1em}{.1em}
			& DTU Scan        & 37    & 97    & 24    & 40    & 55    & 63    & 65    & 69    & 83    & 105   & 106   & 110   & 114   & 118   & 122   & Mean  \\ \hline
			\multirow{7}{*}{\begin{tabular}[c]{@{}l@{}}Novel View\\ Render\\ \\ PSNR\end{tabular}} & Nerfactor \cite{zhang2020nerfactor}      & 21.91 & 20.45 & 23.24 & 23.33 & 26.86 & 22.70 & 24.71 & 27.59 & 22.56 & 25.08 & 26.30 & 25.14 & 21.35 & 26.44 & 26.53 & 24.28 \\
            & PhySG \cite{zhang2020physg} & 15.11 & 17.31 & 17.38 & 20.65 & 18.71 & 18.89 & 18.47 & 18.08 & 21.98 & 20.67 & 18.75 & 17.55 & 21.20 & 18.78 & 23.16 & 19.11 \\
            & Neural-PIL \cite{boss2021neuralpil} & 19.51 & 19.88 & 20.67 & 19.12 & 21.01 & 23.70 & 18.94 & 17.05 & 20.54 & 19.67 & 18.20 & 17.75 & 21.38 & 21.69 & - & 19.94 \\
			     & Ours (MLP-Sep.) & 23.56 & 23.85 & 26.71 & \textbf{28.30} & 29.19 & 28.15 & 26.96 & 29.86 & 25.29 & 27.50 & \textbf{31.88} & 28.43 & 27.16 & 31.52 & 33.30 & 28.11 \\
			       & Ours (MLP)      & 24.17 & \textbf{24.60} & 26.40 & 27.24 & 29.85 & \textbf{28.16} & \textbf{27.39} & 29.82 & \textbf{25.50} & 28.19 & 31.84 & 30.20 & \textbf{27.71} & 30.87 & \textbf{33.62} & 28.37 \\
			& Ours (NGP-Sep.) & 26.00 & 24.27 & 27.14 & 28.09 & \textbf{30.22} & 27.54 & 26.78 & 30.72 & 24.44 & 29.19 & 31.63 & 30.48 & 26.88 & \textbf{31.58} & 32.86 & 28.52 \\
			& Ours (NGP)      & \textbf{26.21} & 24.56 & \textbf{27.31} & 28.19 & 30.07 & 27.47 & 26.79 & \textbf{30.92} & 24.63 & \textbf{29.25} & 31.58 & \textbf{30.69} & 26.93 & 31.33 & 33.19 & \textbf{28.61} \\ \hline
			\multirow{7}{*}{\begin{tabular}[c]{@{}l@{}}Geometry\\ \\ Chamfer\\ Distance\\ (mm)\end{tabular}}           & NeRF  \cite{mildenhall2020nerf}          & 1.920 & 1.730 & 1.920 & 0.800 & 3.410 & 1.390 & 1.510 & 5.440 & 2.040 & 1.100 & 1.010 & 2.880 & 0.910 & 1.000 & 0.790 & 1.857 \\
			    & VolSDF  \cite{yariv2021volume}        & 1.140 & 1.260 & 0.810 & 0.490 & 1.250 & 0.700 & 0.720 & 1.290 & 1.180 & \textbf{0.700} & 0.660 & 1.080 & 0.420 & 0.610 & 0.550 & 0.857 \\
			   & NeuS  \cite{wang2021neus}          & 1.000 & 1.370 & 0.930 & \textbf{0.430} & 1.100 & \textbf{0.650} & \textbf{0.570} & 1.480 & \textbf{1.090} & 0.830 & \textbf{0.520} & 1.200 & \textbf{0.350} & 0.490 & 0.540 & 0.837 \\
			& Ours (MLP-Sep.) & 1.593 & 1.998 & 0.919 & 0.498 & 1.116 & 0.899 & 0.813 & 1.509 & 1.274 & 1.109 & 0.723 & 1.852 & 0.428 & 0.654 & 0.653 & 1.069 \\
			& Ours (MLP)      & 1.303 & 1.911 & 0.954 & 0.595 & 1.271 & 0.881 & 0.860 & 1.529 & 1.242 & 1.125 & 0.723 & 2.096 & 0.488 & 0.838 & 0.654 & 1.098 \\
			& Ours (NGP-Sep.) & 0.608 & \textbf{1.017} & 0.470 & 0.466 & 0.701 & 0.777 & 0.784 & \textbf{1.266} & 1.117 & 0.841 & 0.606 & \textbf{0.926} & 0.388 & \textbf{0.552} & 0.506 & \textbf{0.735} \\
			& Ours (NGP)      & \textbf{0.602} & 1.034 & \textbf{0.466} & 0.470 & \textbf{0.699} & 0.773 & 0.783 & 1.281 & 1.125 & 0.848 & 0.614 & 0.936 & 0.398 & 0.559 & \textbf{0.501} & 0.739 \\ \hline
			\multirow{4}{*}{\begin{tabular}[c]{@{}l@{}}Normal\\ \\ Angle ($^\circ$)\end{tabular}}    &     Ours (MLP-Sep.)          & 26.49 & 28.11 & 14.04 & 15.88 & 18.58 & 14.57 & 22.10 & 31.90 & 30.65 & 34.00 & 15.77 & 25.81 & 16.40 & 17.74 & 18.71 & 22.05 \\
			      & Ours (MLP)      & 24.55 & 26.79 & 13.82 & 15.67 & 18.21 & 14.08 & 21.94 & 31.62 & \textbf{29.36} & 34.07 & 15.96 & 25.64 & 16.90 & 19.80 & 18.85 & 21.82 \\
			& Ours (NGP-Sep.) & 19.98 & 24.27 & 12.10 & \textbf{13.31} & 15.50 & \textbf{13.82} & 18.98 & 29.87 & 31.23 & 32.55 & \textbf{14.47} & 21.66 & \textbf{16.01} & \textbf{14.66} & \textbf{15.88} & 19.62 \\
			& Ours (NGP)      & \textbf{19.61} & \textbf{23.81} & \textbf{11.85} & 13.40 & \textbf{15.40} & 13.86 & \textbf{18.91} & \textbf{29.56} & 29.93 & \textbf{32.37} & 14.49 & \textbf{21.50} & 16.06 & 14.69 & 15.90 & \textbf{19.42} \\
			\specialrule{.2em}{.1em}{.1em}
	\end{tabular}}
	\caption{Quantitative results of geometry accuracy and novel view rendering quality. The proposed method outperforms previous methods in terms of novel view rendering quality and geometry accuracy. For the ablation study on the joint optimization, the full system achieves better novel view rendering quality, and does not degrade geometry accuracy overall. }
	\label{tab:quantitative}
\end{table*}

\subsection{Evaluation on Joint Optimization}\label{sec:exp-joint}
We do an ablation study on the effect of joint optimization on geometry quality. The experiments are conducted on the DTU dataset. Although the images are in LDR and thus the estimated material is not reliable, we consider it a suitable benchmarking dataset for geometry because it has been extensively tested by previous methods. We report the Chamfer distance between the exported mesh and the ground truth. Additionally, we evaluate the normal difference of the nearest points during the calculation of Chamfer distance like \cite{zhang2022critical}. The reason is that one of the main contributions of surface refinement is the normal optimization from the physically-based rendering, which is similar to Shape-from-Shading. However, the vanilla Chamfer distance metric is not sufficient to detect the improvement of surface normal because the point cloud sampled from a bumpy surface can also have an overall small nearest neighbor distance. The additional normal metric can indicate the accuracy of geometry more comprehensively. 

As shown in Fig.~\ref{fig:comparison}, the joint refinement improves both the geometry and material estimation quality. Dumps and dents resulted from the shape-radiance ambiguity can be smoothed by joint refinement. And it can also mitigate the case that the discontinuity of base color is wrongly imprinted on geometry. Quantitative results are shown in Tab.~\ref{tab:quantitative} (Ours (MLP-Sep.) v.s. Ours (MLP) and Ours (NGP-Sep.) v.s. Ours (NGP)). Although the Chamfer distance of the jointly trained system is slightly worse than the separately trained one, the normal consistency is better, which is as expected. As a result, the quality of novel view rendering is also better. In summary, the refinement effect can be detected qualitatively, but is not significant enough in quantitative evaluations. 

\begin{figure*}
	\centering
	\includegraphics[width=1\linewidth]{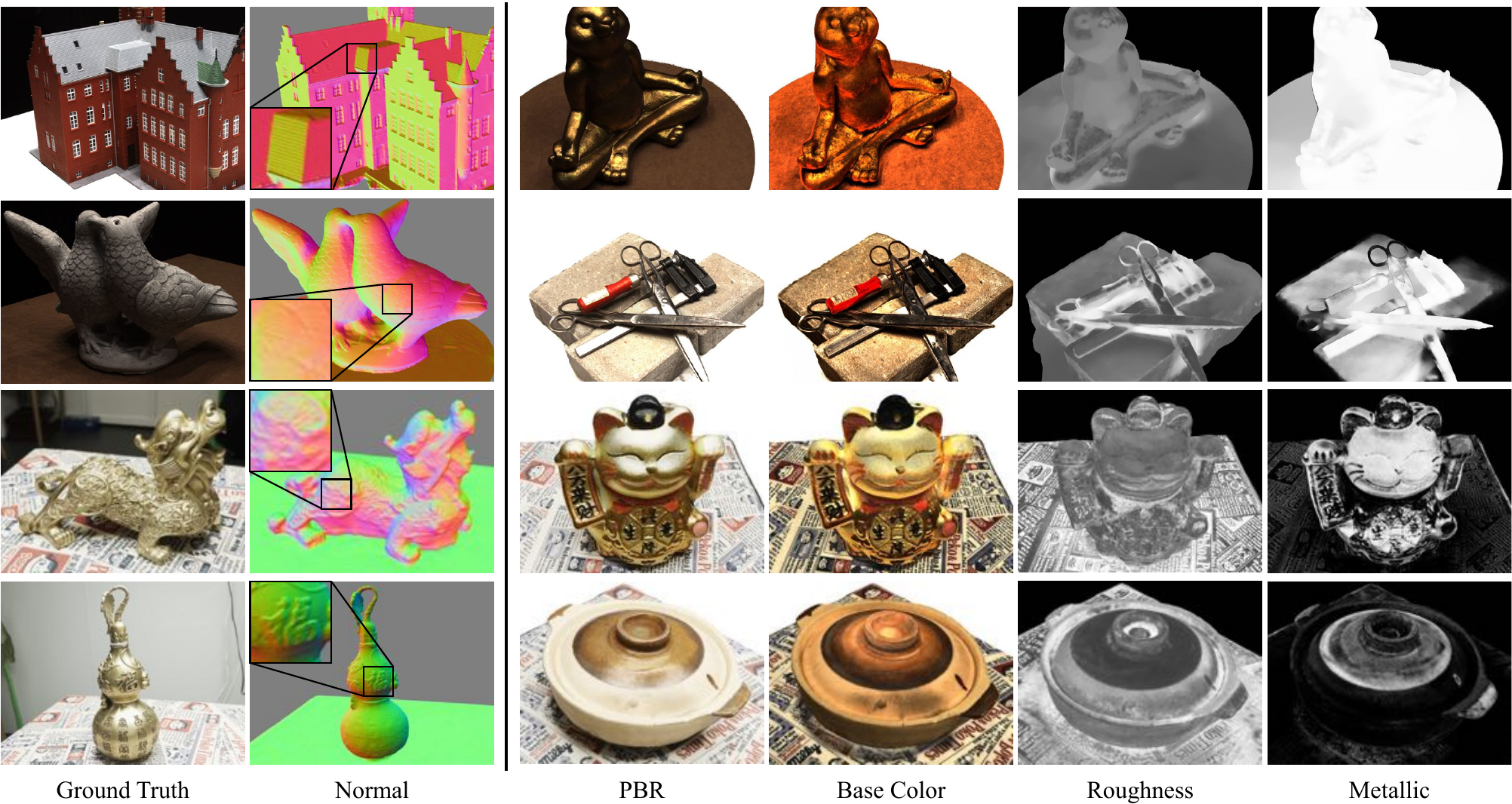}
	\caption{Qualitative geometry and material estimation results on DTU \cite{jensen2014large} and NeILF-HDR. The proposed method can generate surface with fine details and reasonable material. }
	\label{fig:qual}
    \vspace{-3mm}
\end{figure*}

\begin{figure}
	\centering
	\includegraphics[width=1\linewidth]{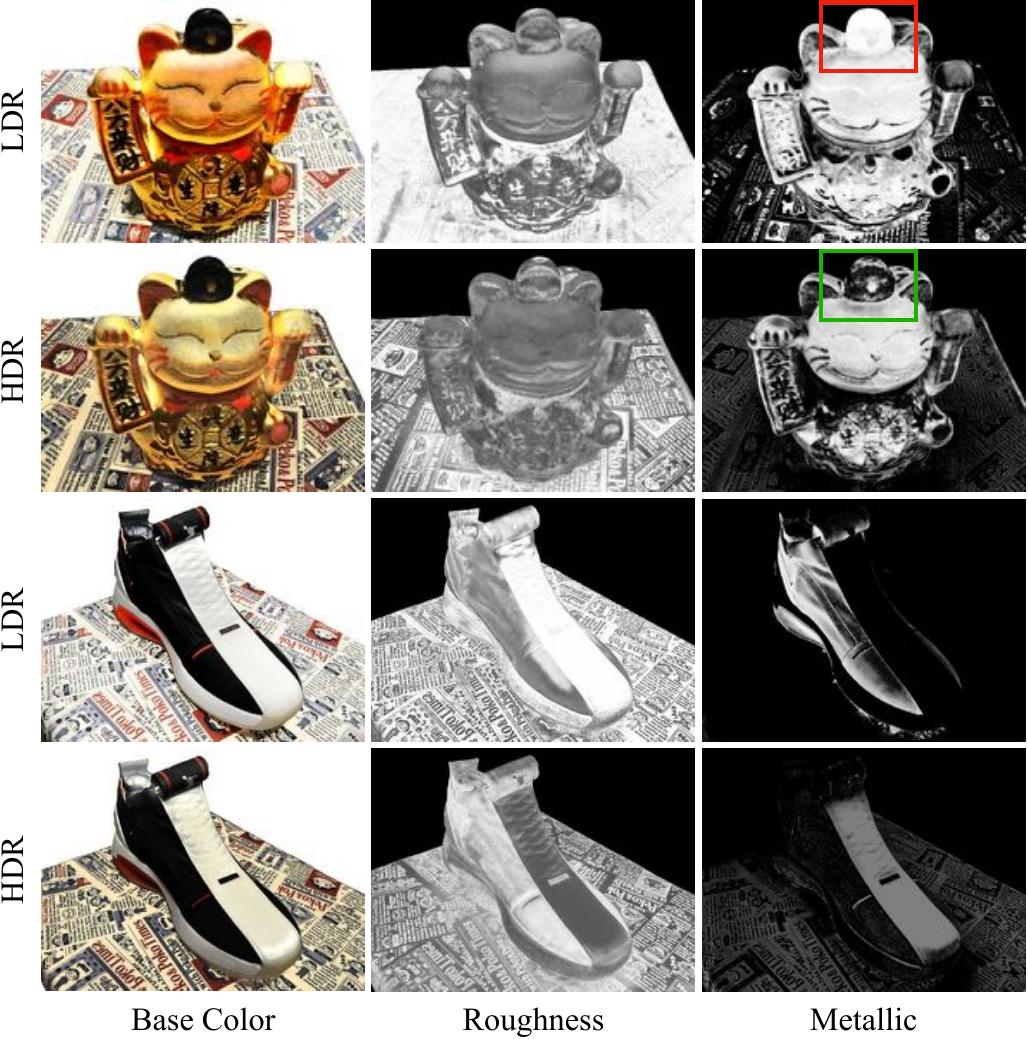}
	\caption{Comparisons of estimated material from LDR and HDR images. The results from HDR inputs are more reasonable. }
	\label{fig:hdr}
    \vspace{-3mm}
\end{figure}

\subsection{Geometry and Material Estimation}\label{sec:exp-overall}

We compare our method with previous neural implicit reconstruction methods and material estimation methods. Quantitative results on DTU are shown in Tab.~\ref{tab:quantitative}. Our method (Ours (NGP)) outperforms previous methods on both geometry accuracy and rendering fidelity. And the qualitative geometry and material estimation results are illustrated in Fig.~\ref{fig:qual}. Our method can recover fine details of the surfaces and generate reasonable material. 

\fakepara{NeILF-HDR}
We present the NeILF-HDR dataset containing 10 scenes with multi-view linear HDR images. We capture around 100 views for each scene with a handheld camera, and the trajectory forms 3 loops with different altitudes. The scene is lit by a circular lamp above the object as well as dim ambient light. This setup is mainly for preventing the cameraman from blocking the light sources. For the color space processing, We first capture raw images, do basic denoising steps and exposure adjustment so that all images have RGB channels, and store actual radiance intensity up to a global scale. 

The qualitative geometry and material results on NeILF-HDR dataset are also shown in Fig.~\ref{fig:qual}. 
Additionally, we compare the material estimated from LDR and HDR images respectively and qualitative results are shown in Fig.~\ref{fig:hdr}. We find that the material in the region with extreme base color is often erroneous if supervised with LDR inputs, while the system trained by HDR images shows better robustness, which shows the necessity of HDR inputs. 


\section{Limitations and Future Works}
\fakepara{Input requirements}
It still requires careful photo capturing for our system to produce a decent estimation. For example, we expect the views to be dense enough to properly reconstruct the surface. Also in Sec.~\ref{sec:hdr} we discuss the necessity of HDR inputs. Moreover, we need to guarantee that the lighting condition remains fixed during the capturing process. Although we already relax the requirement of dedicated capturing devices as in some previous works, it is still tedious for non-professional users. 

\fakepara{Robustness}
The solution space for the joint optimization of 4 fields is very large. In this work, we facilitate the convergence by pretraining each component and introducing regularizations. The former costs extra time and the latter may drive the solution to a suboptimal state. In the future, we may use learned geometry and material prior to bootstrap the optimization. 

\fakepara{Model translucency}
In this work we focus on opaque objects: we use SDF as geometry representation and BRDF as surface material. However, light field estimation for translucent objects is also worth investigating. In future work, the proposed work can be extended to support non-opaque geometry and model transmittance distribution function. 

\section{Conclusion}

We have presented a joint geometry, material, and lighting estimation system from multi-view images, among which the light field consists of both outgoing and incident light fields. The two light fields can be unified by an inter-reflection constraint that the outgoing radiance and the incident light of the two ends of the line segments between surfaces should be consistent. 
On the one hand, the introduction of the outgoing radiance field and the inter-refection constraint regularizes the incident light and thus improves the estimated material. On the other hand, the geometry obtained from outgoing light optimization is also refined during the joint training with the incident light and material. 
The proposed method has been extensively evaluated on our in-house synthetic dataset, the real-world DTU dataset, and an in-house captured linear HDR dataset NeILF-HDR. Our method outperforms previous methods in terms of geometry accuracy and novel view rendering quality. 

\clearpage

{\small
\bibliographystyle{ieee_fullname}
\bibliography{egbib}

\begin{thebibliography}{10}\itemsep=-1pt

\bibitem{azinovic2019inverse}
Dejan Azinovic, Tzu-Mao Li, Anton Kaplanyan, and Matthias Nie{\ss}ner.
\newblock Inverse path tracing for joint material and lighting estimation.
\newblock In {\em CVPR}, 2019.

\bibitem{bi2020deep}
Sai Bi, Zexiang Xu, Kalyan Sunkavalli, Milo{\v{s}} Ha{\v{s}}an, Yannick
  Hold-Geoffroy, David Kriegman, and Ravi Ramamoorthi.
\newblock Deep reflectance volumes: Relightable reconstructions from multi-view
  photometric images.
\newblock In {\em ECCV}, 2020.

\bibitem{bi2020deep3d}
Sai Bi, Zexiang Xu, Kalyan Sunkavalli, David Kriegman, and Ravi Ramamoorthi.
\newblock Deep 3d capture: Geometry and reflectance from sparse multi-view
  images.
\newblock In {\em CVPR}, 2020.

\bibitem{boss2021nerd}
Mark Boss, Raphael Braun, Varun Jampani, Jonathan~T Barron, Ce Liu, and Hendrik
  Lensch.
\newblock Nerd: Neural reflectance decomposition from image collections.
\newblock In {\em ICCV}, 2021.

\bibitem{boss2021neuralpil}
Mark Boss, Varun Jampani, Raphael Braun, Ce Liu, Jonathan Barron, and
  Hendrik~PA Lensch.
\newblock Neural-pil: Neural pre-integrated lighting for reflectance
  decomposition.
\newblock In {\em NeurIPS}, 2021.

\bibitem{blender}
Blender~Online Community.
\newblock {\em Blender - a 3D modelling and rendering package}.
\newblock Blender Foundation, Stichting Blender Foundation, Amsterdam, 2018.

\bibitem{dong2014appearance}
Yue Dong, Guojun Chen, Pieter Peers, Jiawan Zhang, and Xin Tong.
\newblock Appearance-from-motion: Recovering spatially varying surface
  reflectance under unknown lighting.
\newblock {\em TOG}, 2014.

\bibitem{guo2019relightables}
Kaiwen Guo, Peter Lincoln, Philip Davidson, Jay Busch, Xueming Yu, Matt Whalen,
  Geoff Harvey, Sergio Orts-Escolano, Rohit Pandey, Jason Dourgarian, et~al.
\newblock The relightables: Volumetric performance capture of humans with
  realistic relighting.
\newblock {\em TOG}, 2019.

\bibitem{jensen2014large}
Rasmus Jensen, Anders Dahl, George Vogiatzis, Engil Tola, and Henrik Aan{\ae}s.
\newblock Large scale multi-view stereopsis evaluation.
\newblock In {\em CVPR}, 2014.

\bibitem{kajiya1986rendering}
James~T Kajiya.
\newblock The rendering equation.
\newblock In {\em Proceedings of the 13th annual conference on Computer
  graphics and interactive techniques}, 1986.

\bibitem{li2022neural}
Junxuan Li and Hongdong Li.
\newblock Neural reflectance for shape recovery with shadow handling.
\newblock In {\em CVPR}, 2022.

\bibitem{ling2022shadowneus}
Jingwang Ling, Zhibo Wang, and Feng Xu.
\newblock Shadowneus: Neural sdf reconstruction by shadow ray supervision.
\newblock {\em arXiv preprint arXiv:2211.14086}, 2022.

\bibitem{martin2021nerf}
Ricardo Martin-Brualla, Noha Radwan, Mehdi~SM Sajjadi, Jonathan~T Barron,
  Alexey Dosovitskiy, and Daniel Duckworth.
\newblock Nerf in the wild: Neural radiance fields for unconstrained photo
  collections.
\newblock In {\em CVPR}, 2021.

\bibitem{mildenhall2020nerf}
Ben Mildenhall, Pratul~P Srinivasan, Matthew Tancik, Jonathan~T Barron, Ravi
  Ramamoorthi, and Ren Ng.
\newblock Nerf: Representing scenes as neural radiance fields for view
  synthesis.
\newblock In {\em ECCV}, 2020.

\bibitem{mueller2022instant}
Thomas M\"uller, Alex Evans, Christoph Schied, and Alexander Keller.
\newblock Instant neural graphics primitives with a multiresolution hash
  encoding.
\newblock {\em arXiv:2201.05989}, 2022.

\bibitem{munkberg2021nvdiffrec}
Jacob Munkberg, Jon Hasselgren, Tianchang Shen, Jun Gao, Wenzheng Chen, Alex
  Evans, Thomas Mueller, and Sanja Fidler.
\newblock {Extracting Triangular 3D Models, Materials, and Lighting From
  Images}.
\newblock {\em arXiv:2111.12503}, 2021.

\bibitem{nam2018practical}
Giljoo Nam, Joo~Ho Lee, Diego Gutierrez, and Min~H Kim.
\newblock Practical svbrdf acquisition of 3d objects with unstructured flash
  photography.
\newblock {\em TOG}, 2018.

\bibitem{niemeyer2020differentiable}
Michael Niemeyer, Lars Mescheder, Michael Oechsle, and Andreas Geiger.
\newblock Differentiable volumetric rendering: Learning implicit 3d
  representations without 3d supervision.
\newblock In {\em CVPR}, 2020.

\bibitem{nimier2021material}
Merlin Nimier-David, Zhao Dong, Wenzel Jakob, and Anton Kaplanyan.
\newblock Material and lighting reconstruction for complex indoor scenes with
  texture-space differentiable rendering.
\newblock Eurographics Symposium on Rendering, 2021.

\bibitem{park2020seeing}
Jeong~Joon Park, Aleksander Holynski, and Steven~M Seitz.
\newblock Seeing the world in a bag of chips.
\newblock In {\em CVPR}, 2020.

\bibitem{schmitt2020joint}
Carolin Schmitt, Simon Donne, Gernot Riegler, Vladlen Koltun, and Andreas
  Geiger.
\newblock On joint estimation of pose, geometry and svbrdf from a handheld
  scanner.
\newblock In {\em CVPR}, 2020.

\bibitem{schonberger2016pixelwise}
Johannes~L Sch{\"o}nberger, Enliang Zheng, Jan-Michael Frahm, and Marc
  Pollefeys.
\newblock Pixelwise view selection for unstructured multi-view stereo.
\newblock In {\em ECCV}, 2016.

\bibitem{srinivasan2021nerv}
Pratul~P Srinivasan, Boyang Deng, Xiuming Zhang, Matthew Tancik, Ben
  Mildenhall, and Jonathan~T Barron.
\newblock Nerv: Neural reflectance and visibility fields for relighting and
  view synthesis.
\newblock In {\em CVPR}, 2021.

\bibitem{walter2007microfacet}
Bruce Walter, Stephen~R Marschner, Hongsong Li, and Kenneth~E Torrance.
\newblock Microfacet models for refraction through rough surfaces.
\newblock In {\em Proceedings of the 18th Eurographics conference on Rendering
  Techniques}, pages 195--206, 2007.

\bibitem{wang2009all}
Jiaping Wang, Peiran Ren, Minmin Gong, John Snyder, and Baining Guo.
\newblock All-frequency rendering of dynamic, spatially-varying reflectance.
\newblock In {\em ACM SIGGRAPH Asia}, 2009.

\bibitem{wang2021neus}
Peng Wang, Lingjie Liu, Yuan Liu, Christian Theobalt, Taku Komura, and Wenping
  Wang.
\newblock Neus: Learning neural implicit surfaces by volume rendering for
  multi-view reconstruction.
\newblock 2021.

\bibitem{xia2016recovering}
Rui Xia, Yue Dong, Pieter Peers, and Xin Tong.
\newblock Recovering shape and spatially-varying surface reflectance under
  unknown illumination.
\newblock {\em TOG}, 2016.

\bibitem{yao2018mvsnet}
Yao Yao, Zixin Luo, Shiwei Li, Tian Fang, and Long Quan.
\newblock Mvsnet: Depth inference for unstructured multi-view stereo.
\newblock In {\em ECCV}, 2018.

\bibitem{yao2022neilf}
Yao Yao, Jingyang Zhang, Jingbo Liu, Yihang Qu, Tian Fang, David McKinnon,
  Yanghai Tsin, and Long Quan.
\newblock Neilf: Neural incident light field for physically-based material
  estimation.
\newblock In {\em ECCV}, 2022.

\bibitem{yariv2021volume}
Lior Yariv, Jiatao Gu, Yoni Kasten, and Yaron Lipman.
\newblock Volume rendering of neural implicit surfaces.
\newblock {\em arXiv preprint arXiv:2106.12052}, 2021.

\bibitem{yariv2020multiview}
Lior Yariv, Yoni Kasten, Dror Moran, Meirav Galun, Matan Atzmon, Basri Ronen,
  and Yaron Lipman.
\newblock Multiview neural surface reconstruction by disentangling geometry and
  appearance.
\newblock {\em NeurIPS}, 2020.

\bibitem{zhang2022critical}
Jingyang Zhang, Yao Yao, Shiwei Li, Tian Fang, David McKinnon, Yanghai Tsin,
  and Long Quan.
\newblock Critical regularizations for neural surface reconstruction in the
  wild.
\newblock In {\em CVPR}, 2022.

\bibitem{zhang2020visibility}
Jingyang Zhang, Yao Yao, Shiwei Li, Zixin Luo, and Tian Fang.
\newblock Visibility-aware multi-view stereo network.
\newblock {\em BMVC}, 2020.

\bibitem{zhang2021learning}
Jingyang Zhang, Yao Yao, and Long Quan.
\newblock Learning signed distance field for multi-view surface reconstruction.
\newblock In {\em ICCV}, 2021.

\bibitem{zhang2020physg}
Kai Zhang, Fujun Luan, Qianqian Wang, Kavita Bala, and Noah Snavely.
\newblock Physg: Inverse rendering with spherical gaussians for physics-based
  material editing and relighting.
\newblock In {\em CVPR}, 2021.

\bibitem{zhang2020nerf++}
Kai Zhang, Gernot Riegler, Noah Snavely, and Vladlen Koltun.
\newblock Nerf++: Analyzing and improving neural radiance fields.
\newblock {\em arXiv preprint arXiv:2010.07492}, 2020.

\bibitem{zhang2020nerfactor}
Xiuming Zhang, Pratul~P Srinivasan, Boyang Deng, Paul Debevec, William~T
  Freeman, and Jonathan~T Barron.
\newblock {NeRFactor: Neural Factorization of Shape and Reflectance Under an
  Unknown Illumination}.
\newblock {\em TOG}, 2021.

\end{thebibliography}
}

\clearpage

\begin{center}
    \vspace{2mm}
    \textbf{\LARGE Supplementary Material}
\end{center}
\setcounter{section}{0}
\setcounter{equation}{0}
\setcounter{figure}{0}
\setcounter{table}{0}
\setcounter{page}{1}

\vspace{2mm}

\section{Method}
\subsection{BRDF Parameterization}
In Sec.~3.1 we introduce the $D$, $F$, and $G$ term of the specualr BRDF. The normal dsitribution term $D$ is approximated by Spherical Gaussian whose sharpness is controlled by the roughness $r$:
\begin{equation}
    D(\mathbf{h};r) = S(\mathbf{h}, \frac{1}{\pi r^4}, \mathbf{n}, \frac{2}{r^4}) = \frac{1}{\pi r^4} \exp ({\frac{2}{r^4}(\mathbf{h}\cdot \mathbf{n} - 1)}).
\end{equation}
The Fresnel term $F$ is given by:
\begin{equation}
\begin{aligned}
    F(\boldsymbol\omega_o, \mathbf{h}; \mathbf{b}, m) & = F_0 + (1 - F_0) (1- (\boldsymbol\omega_o \cdot \mathbf{h})^5), \\
    \text{where } F_0 & = 0.04(1-m) + \mathbf{b}m.
\end{aligned}
\end{equation}
The geoemtry term $G$ is modeled by two GGX function \cite{walter2007microfacet}:
\begin{equation}
\begin{aligned}
    G(\boldsymbol\omega_i, \boldsymbol\omega_o, \mathbf{n}; r) &= G_{\text{GGX}}(\boldsymbol\omega_i \cdot \mathbf{n})G_{\text{GGX}}(\boldsymbol\omega_o \cdot \mathbf{n}), \\
    \text{where } G_{\text{GGX}}(z) &= \frac{2z}{ (2-r^2)z + r^2 }.
\end{aligned}
\end{equation}

\subsection{Loss Terms}
In this section we introduce the details of the loss terms besides $L_{\text{vol}}$, $L_{\text{phys}}$ and $L_{\text{ref}}$. 

\fakepara{Eikonal loss}
This loss is adopted by most of the SDF-based systems \cite{yariv2020multiview} It expects the expectation of gradient magnitude is 1:
\begin{equation}
    L_{\text{Eik}} = | \| \nabla \mathbb{G}(\mathbf{x}) \| - 1 |.
\end{equation}

\fakepara{Hessian loss}
This loss discourages the direction of the gradient to change rapidly by minimizing the norm of the Hessian matrix \cite{zhang2022critical}:
\begin{equation}
    L_{\text{Hess}} = \| \mathbf{H} \mathbb{G}(\mathbf{x}) \|_1, 
\end{equation}
where $\|\cdot\|_1$ is the element-wize matrix 1-norm and $\mathbf{H}f$ is the Hessian matrix of the a function $f$. 

\fakepara{Minimal surface loss}
This loss minimize the elastic energy of the surface to produce compact interpolation or extrapolation of the surface for unobserved region \cite{zhang2022critical}. To calculate the area of surface given by an implicit function, we model the surface $\pazocal S$ as a differentiation of the object interior $\pazocal V$:
\begin{equation}
\begin{aligned}
    \pazocal V &= \{ \mathbf{x} \in \Omega | \mathbb{G}(\mathbf{x}) < 0 \}, \\
    \pazocal S &= \partial \pazocal V = \{ \mathbf{x} \in \Omega | \mathbb{G}(\mathbf{x}) = 0 \},
\end{aligned}
\end{equation}
where $\Omega$ is the whole 3D space. And the surface area can be derived as:
\begin{equation}
\begin{aligned}
    \text{volume}(\pazocal V) &= \int_{\Omega} H(\mathbb{G}(\mathbf{x})) d\mathbf{x}, \\
    \text{area}(\pazocal S) &= \int_{\Omega} \| \nabla H(\mathbb{G}(\mathbf{x})) \| \\
    &= \int_{\Omega} \delta( \mathbb{G}(\mathbf{x}) ) \| \nabla \mathbb{G}(\mathbf{x}) \|,
\end{aligned}
\end{equation}
where $H$ is the Heaviside function and $\delta$ is the Dirac function. In practice, the gradient magnitude is 1 so can be omitted, and we use a regularized form of the Dirac function parameterized by a sharpness term $\epsilon$. The loss is given by:
\begin{equation}
    L_{\text{surf}} = \delta_{\epsilon}( \mathbb{G}(\mathbf{x}) ) \text{, where } 
    \delta_{\epsilon}(z) = \frac{\epsilon \pi^{-1}}{\epsilon^2 + z^2}. 
\end{equation}

\fakepara{Point cloud supervision}
We can optionally introduce an oriented point cloud to facilitate the convergence in early stage \cite{zhang2022critical}. The distance values and the gradient directions are fit to the point cloud at the locations of the data points. Let $\mathbf{x}_D$ be the points in a point cloud $D$ with normal $\mathbf{n}_D$, the point cloud loss is given by:
\begin{equation}
    L_{\text{pcd}} = |\mathbb{G}(\mathbf{x}_D)| + (1 - \frac{\mathbf{n}_D \cdot \nabla \mathbb{G}(\mathbf{x}_D)}{\| \nabla \mathbb{G}(\mathbf{x}_D) \|}).
\end{equation}

\begin{figure*}
    \centering
    \includegraphics[width=1\linewidth]{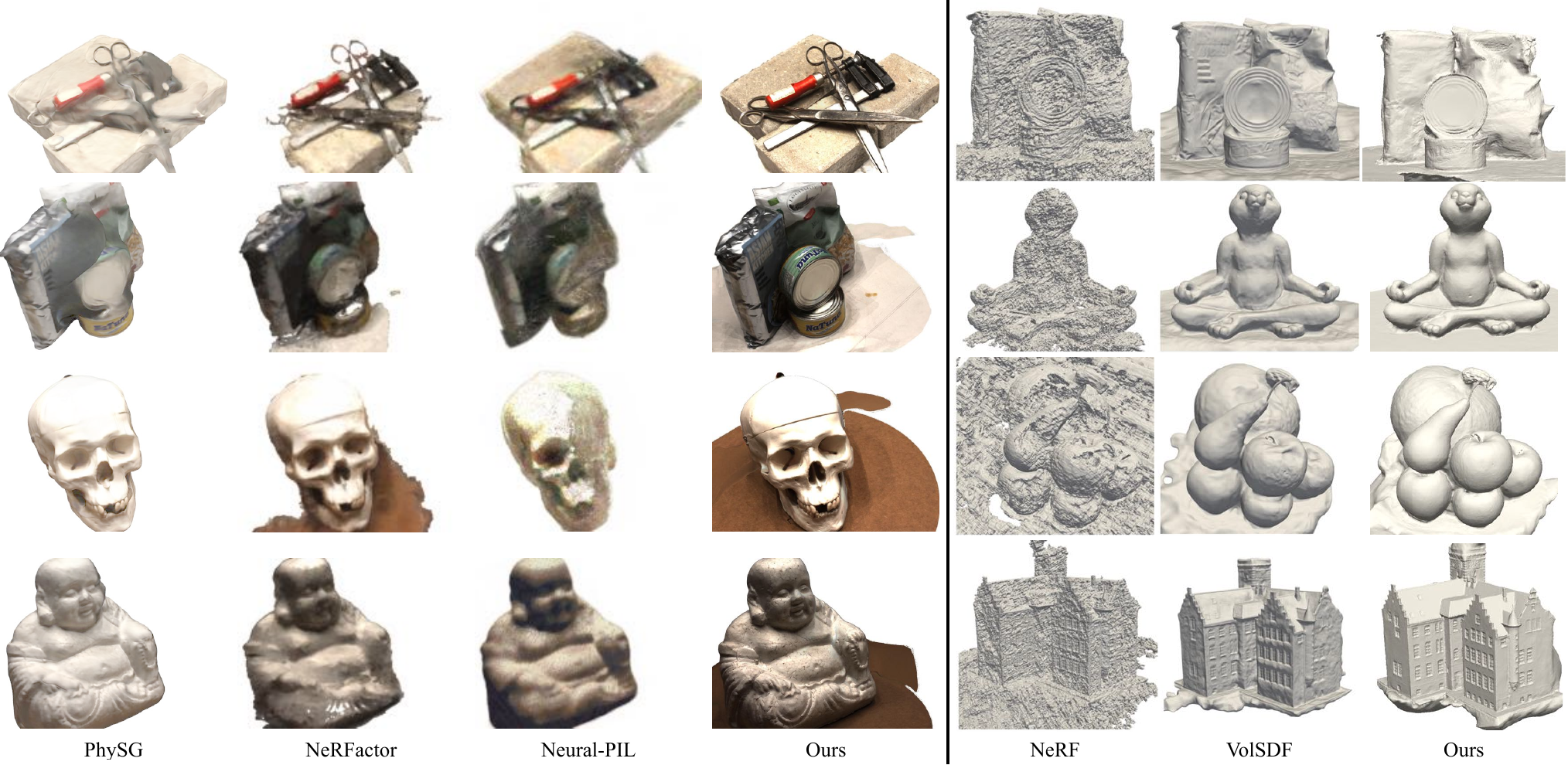}
    \caption{Qualitative comparisons on the DTU \cite{jensen2014large} dataset with previous methods. The geometry results of previous methods are taken from \cite{yariv2021volume}. }
    \label{fig:dtu-compare}
\end{figure*}

\fakepara{Bilateral smoothness}
If the color $\mathbf{I}_{\mathbf{p}}$ of a pixel $\mathbf{p}$ does not change rapidly, we expect that the roughness and the metallic the corresponding 3D surface point $\mathbf{x}_{\mathbf{p}}$ are smooth \cite{yao2022neilf}. The bilateral smoothness loss is given by:
\begin{equation}
    L_{\text{smth}} = ( \|\nabla r(\mathbf{x}_{\mathbf{p}})\| + \|\nabla m(\mathbf{x}_{\mathbf{p}})\| ) \exp(-\|\nabla \mathbf{I}_{\mathbf{p}}\|).
\end{equation}

\fakepara{Lambertian assumption}
We encourage the surface to be Lambertian as a tie break rule during material optimization \cite{yao2022neilf}. The loss is given by:
\begin{equation}
    L_{\text{Lam}} = |r-1| + |m|.
\end{equation}
In practice, the weight for this loss is set to be very small to prevent interfering the material estimation. 

\section{Network Architecture}

\begin{figure}
    \centering
    \includegraphics[width=1\linewidth]{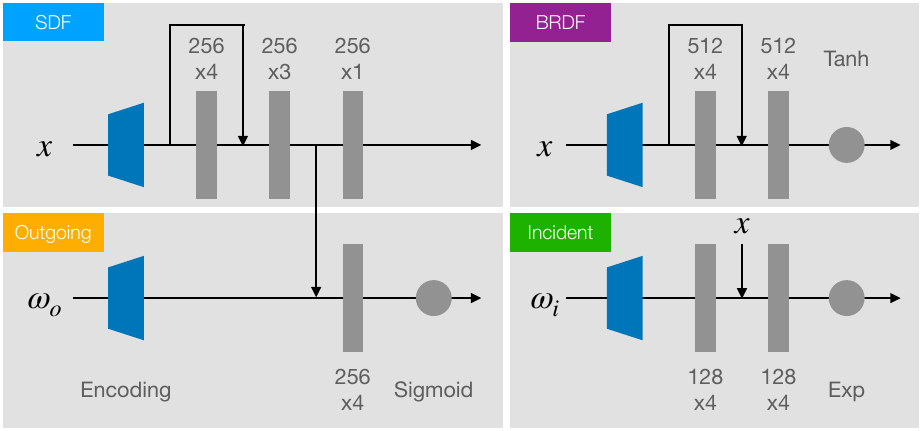}
    \caption{The network architecture for the 4 fields in the \textit{MLP} setting.}
    \label{fig:arch}
\end{figure}

The detailed network architectures for the 4 fields in the \textit{MLP} setting are illustrated in Fig.~\ref{fig:arch}. Note that although the two light fields have the same input, in practice we still separate them because they have different frequency response with respect to each input. The outgoing light field is expected to be more sensitive to the position input. while the incident light field acts in the opposite way. 

The \textit{NGP} setting which uses hash grid \cite{mueller2022instant} as the encoder has less number of layers in general. The SDF, the outgoing radiance field and the BRDF field each has 2, 4, 2 linear layers with 64 hidden units, and the skip connections in the SDF and the BRDF field are removed. 

\section{NeILF-HDR Dataset}

\begin{figure}
    \centering
    \includegraphics[height=0.6\linewidth]{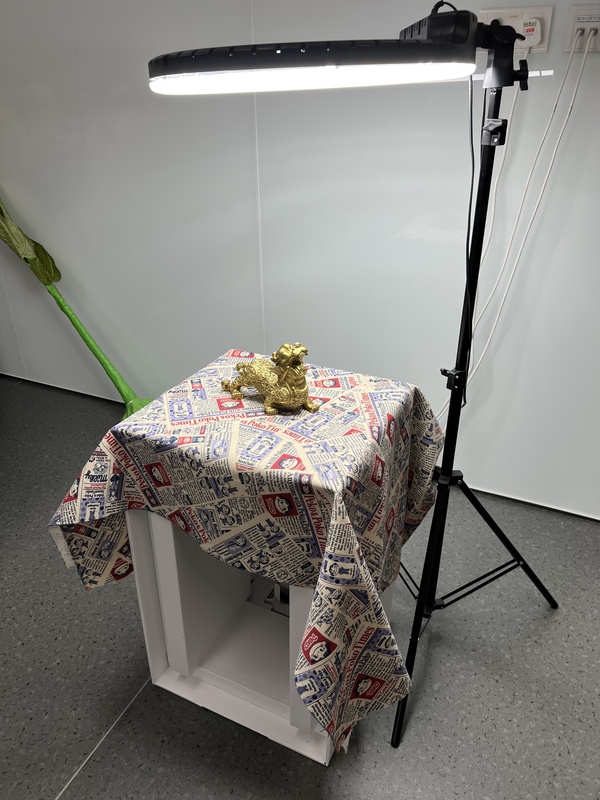}
    \includegraphics[height=0.6\linewidth]{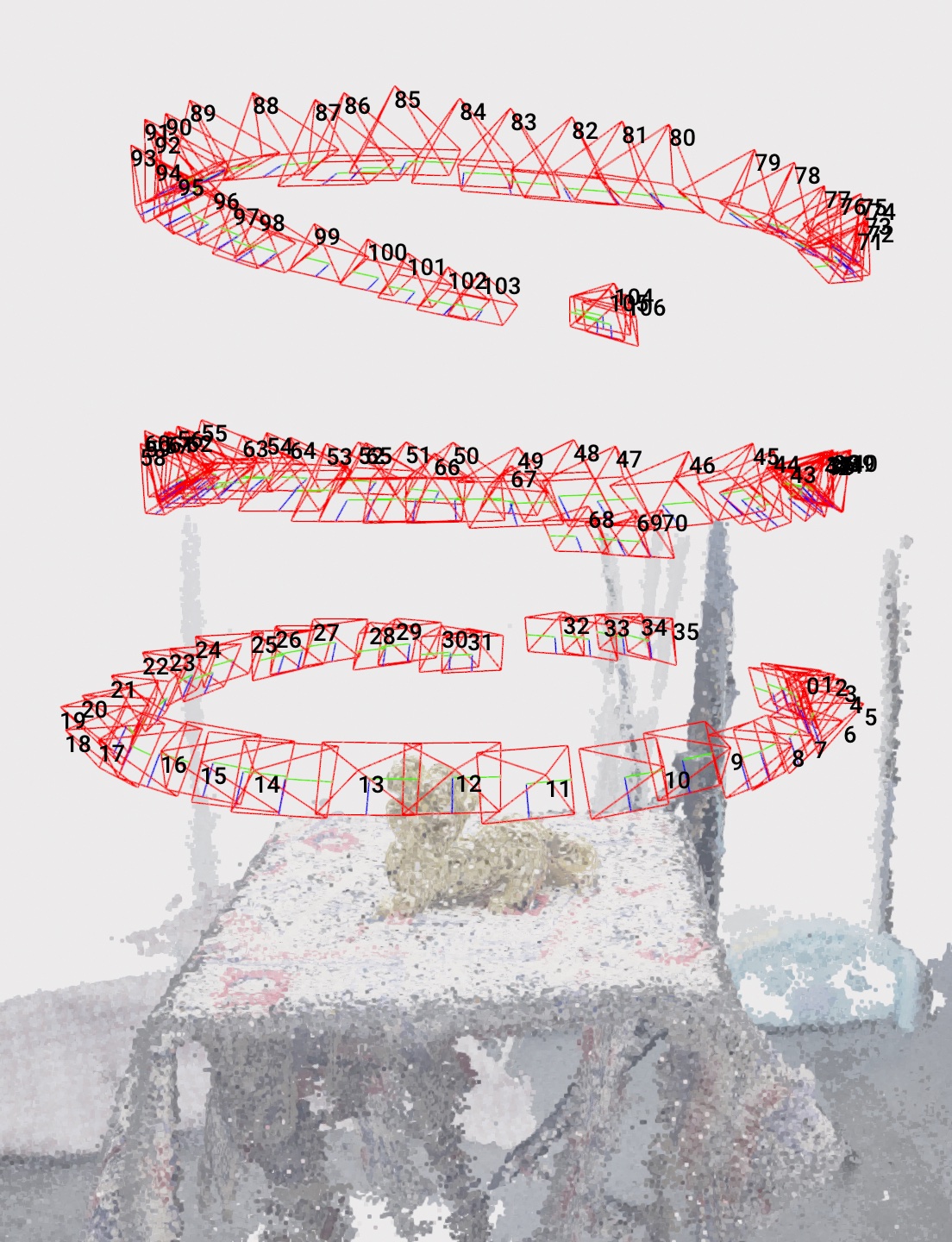}
    \caption{Capturing setup for the NeILF-HDR dataset. The left image shows the lighting setup, and the right image illustrates the camera trajectories.}
    \label{fig:hdr-setup}
\end{figure}

The capturing setup and the camera trajectories are shown in Fig.~\ref{fig:hdr-setup}. 

\begin{figure}
    \centering
    \includegraphics[width=1\linewidth]{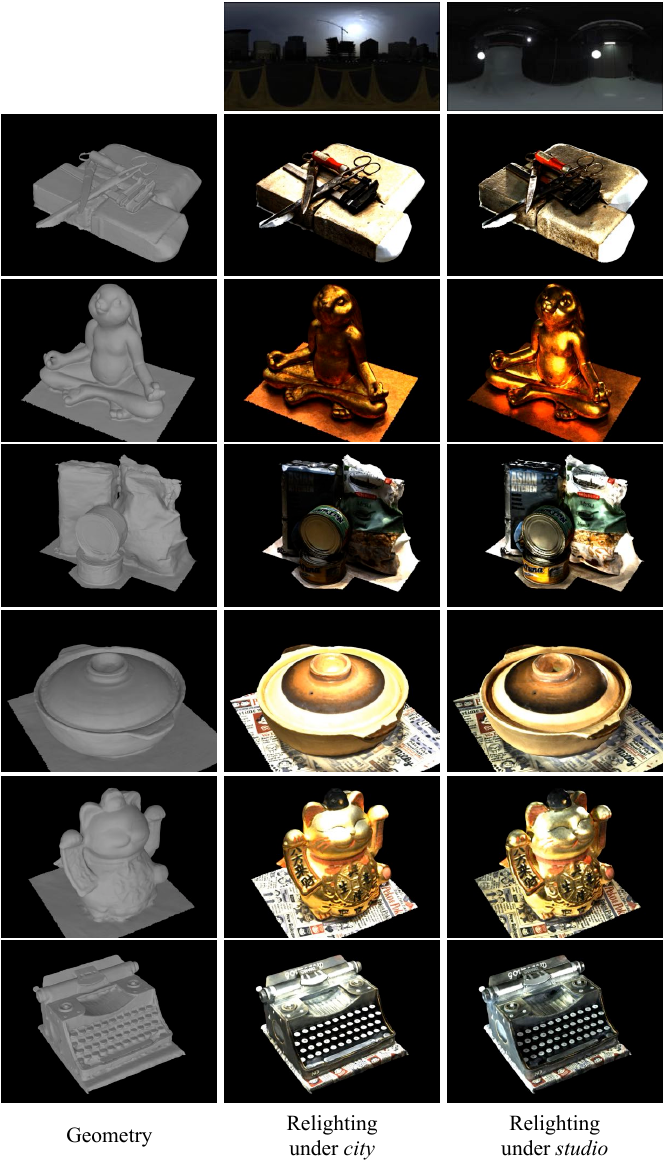}
    \caption{Geometry and relighting results of real world objects. }
    \label{fig:hdr-relight}
\end{figure}

\section{Additional Results}
\subsection{Comparison with Previous Methods}

We qualitatively compare the geometry and novel view rendering of our method with previous methods, and the results are shown in Fig.~\ref{fig:dtu-compare}. 

\subsection{Geometry and Material Estimation}
\begin{figure*}
    \centering
    \includegraphics[width=1\linewidth]{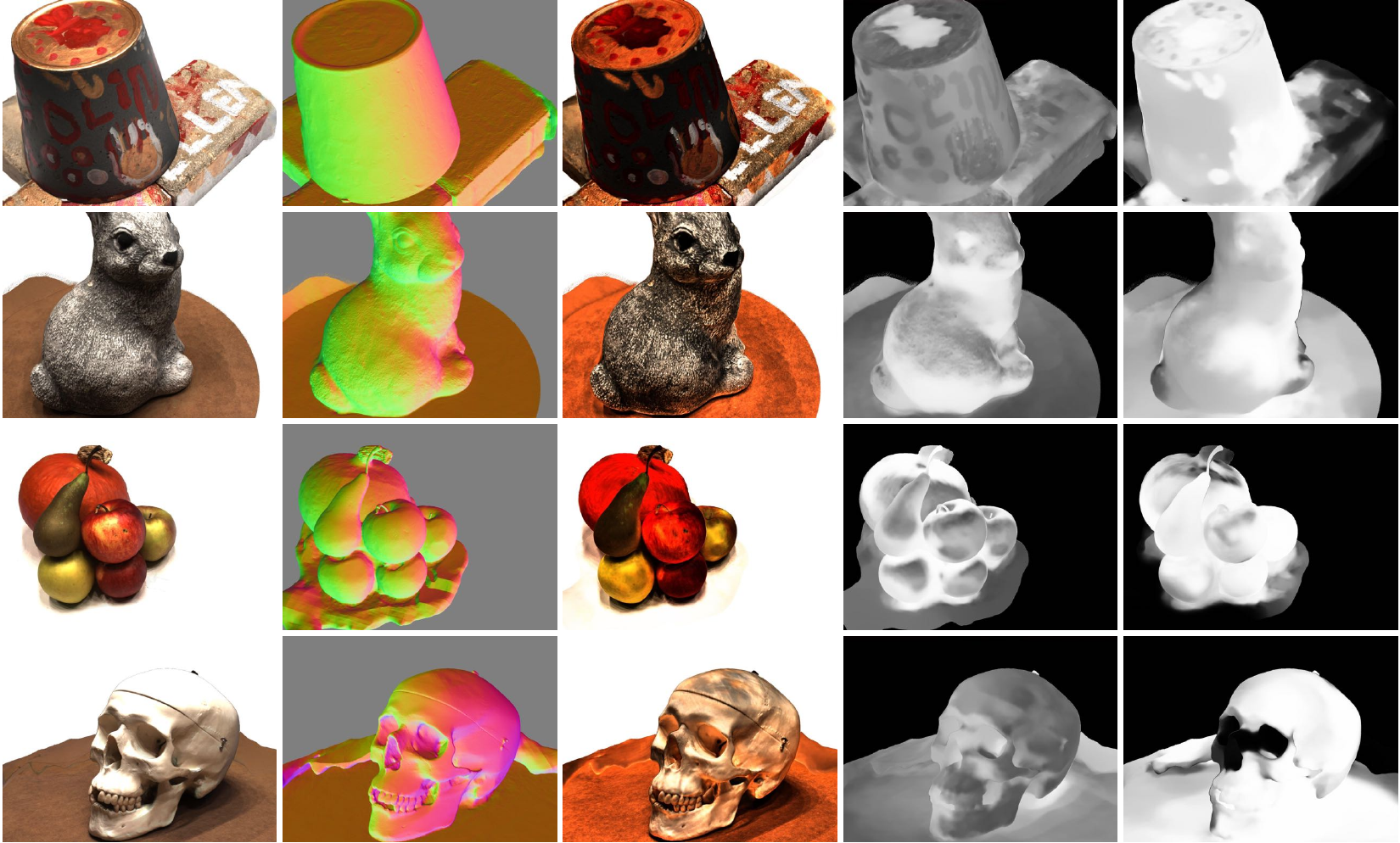}
    \includegraphics[width=1\linewidth]{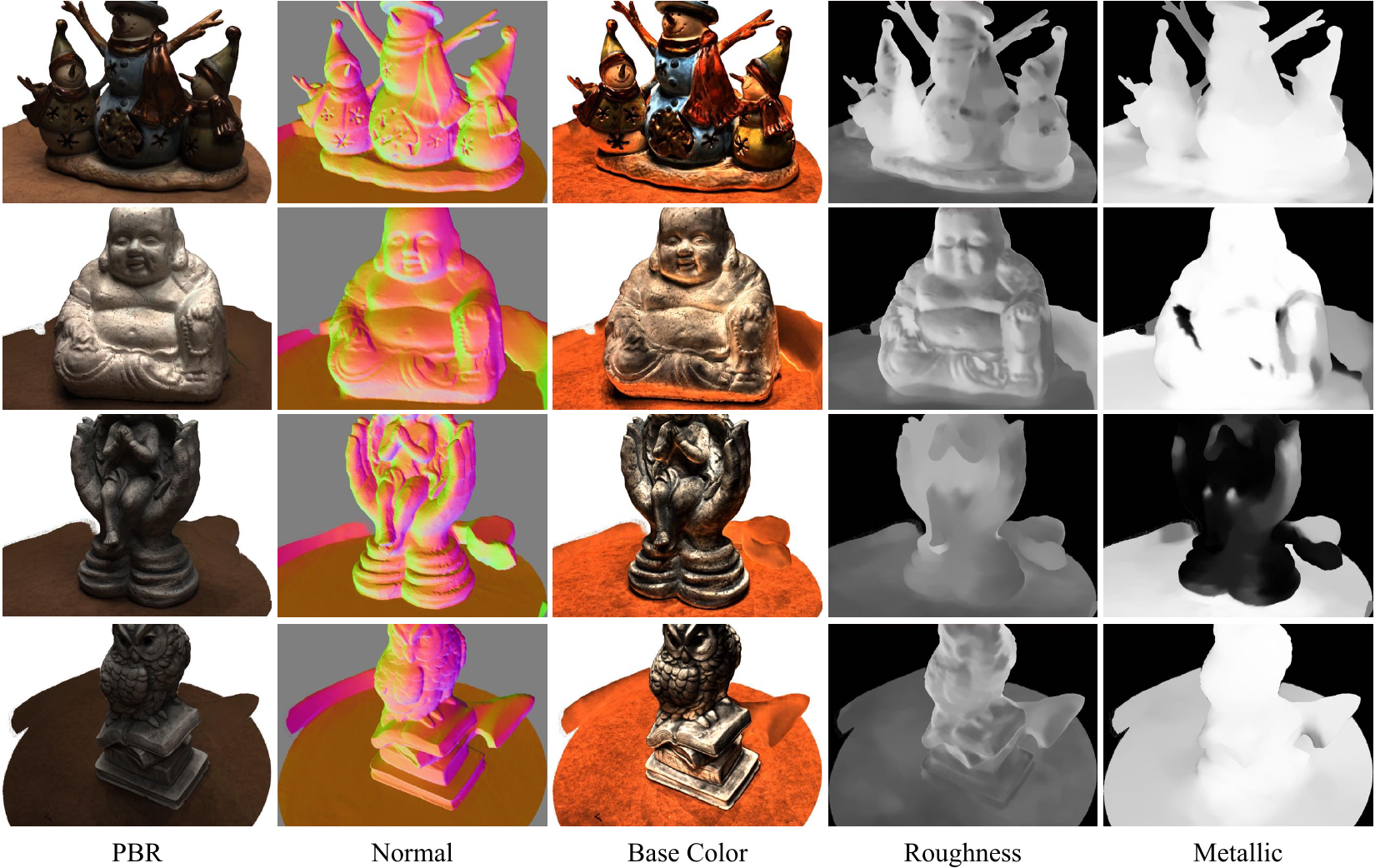}
    \caption{Qualitative results on the DTU \cite{jensen2014large} dataset. The metallic is often overestimated because of LDR input. }
    \label{fig:dtu-qual}
\end{figure*}

\begin{figure*}
    \centering
    \includegraphics[width=1\linewidth]{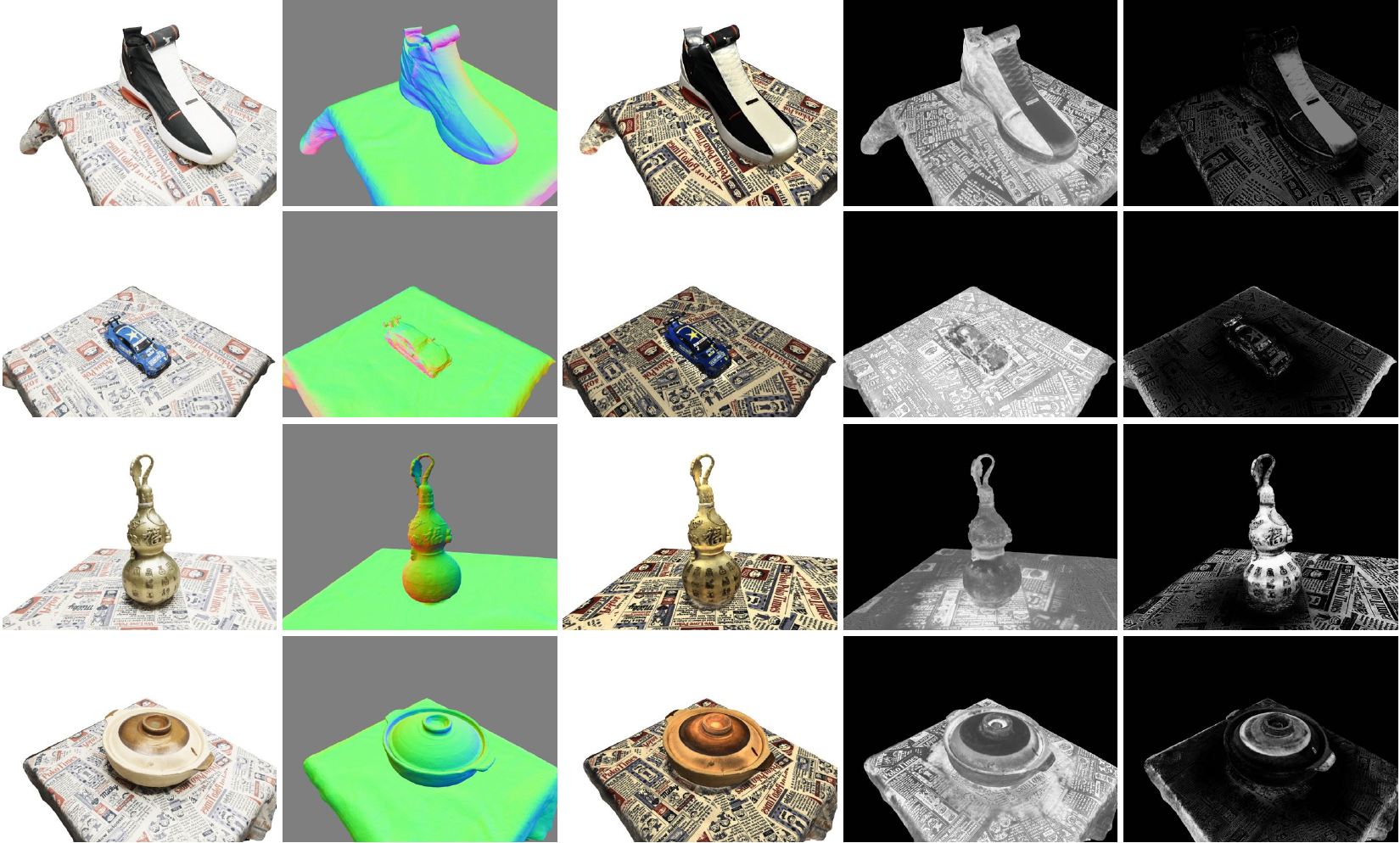}
    \includegraphics[width=1\linewidth]{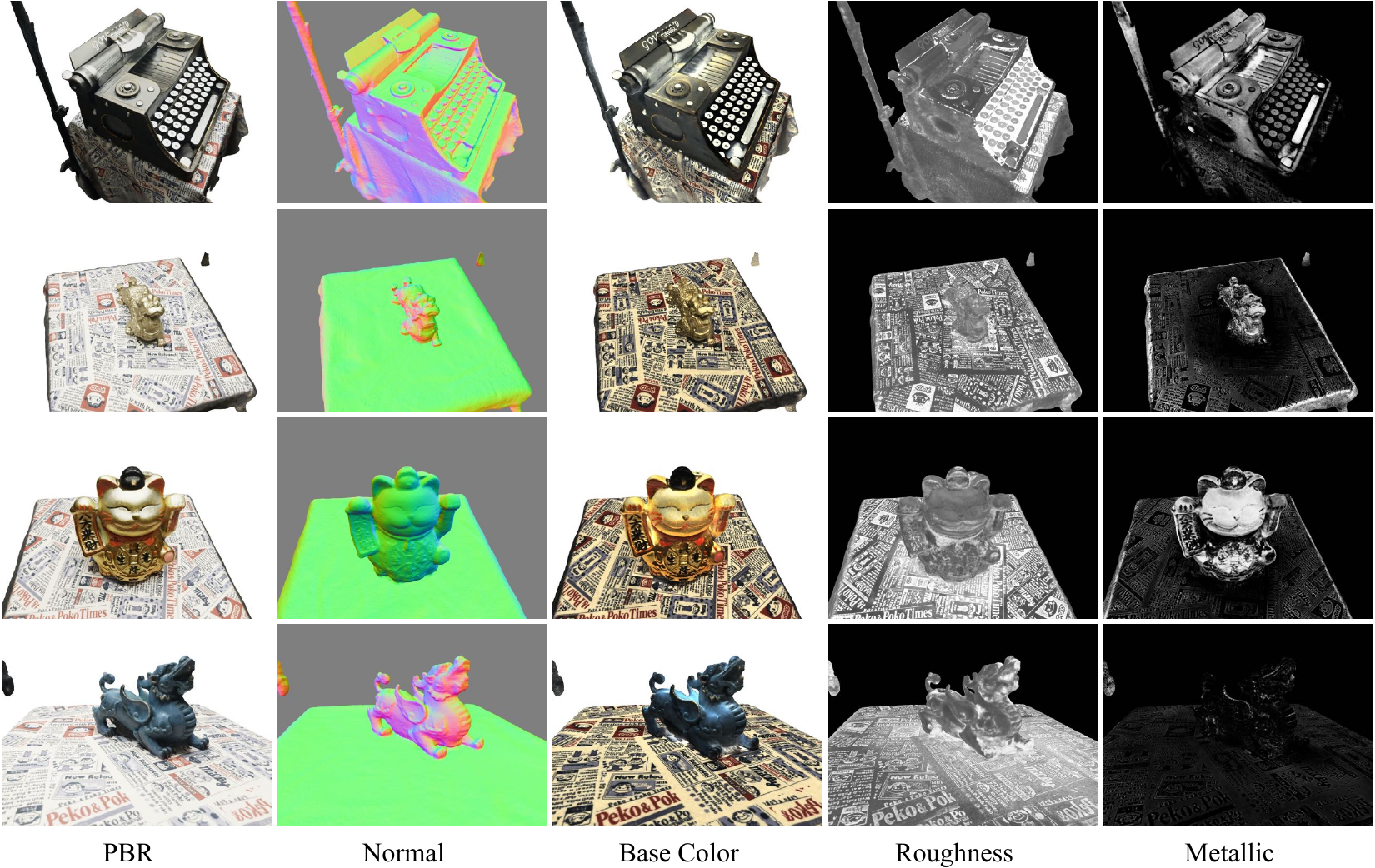}
    \caption{Qualitative results on the NeILF-HDR dataset. }
    \label{fig:hdr-qual}
\end{figure*}

Additional qualitative results of geometry and material estimation results on the DTU \cite{jensen2014large} and the NeILF-HDR dataset are shown in Fig.~\ref{fig:dtu-qual} and \ref{fig:hdr-qual}. Note that the metallic for the DTU scenes is often overestimated because of LDR input. On the NeILF-HDR dataset, the estimated metallic is more reasonable. 

\subsection{Relighting}

We export the material parameters from the BRDF field to a UV map and relight the objects by new environment maps using Blender \cite{blender}. The results are shown in Fig.~\ref{fig:hdr-relight}. For animated results, please refer to the supplementary video. 


\end{document}